\documentclass{article}
\usepackage{spconf,amsmath,graphicx}

\usepackage{comment} 
\usepackage{amsmath}
\usepackage{multirow}
\usepackage{microtype}
\usepackage{graphicx}
\usepackage{xcolor}
\usepackage{hyperref}
\usepackage{makecell}
\usepackage{rotating}
\usepackage{colortbl}
\usepackage{tablefootnote}
\usepackage{xstring}
\usepackage{latexsym}
\usepackage{ragged2e}
\usepackage{linguex}
\usepackage{arabtex}
\usepackage{utf8}
\usepackage{subfig}
\usepackage[utf8]{inputenc}
\usepackage{enumitem}
\setlist[itemize]{noitemsep, nolistsep}

\newcommand*{\MyIndent}{\hspace*{0.3cm}}
\newcommand*{\img}[1]{%

    \raisebox{-.6\baselineskip}{%
        \includegraphics[
        height=2 cm, %\baselineskip
        width=8.6 cm, %\baselineskip,
        ]{#1}%
    }%
}

\title{Textual Data Augmentation for Arabic-English Code-Switching Speech Recognition}
\name{Amir Hussein$^{1,2}$, Shammur Absar Chowdhury$^{3}$, Ahmed Abdelali$^{3}$, Najim Dehak$^{2}$, Ahmed Ali$^{3}$, Sanjeev Khudanpur$^{2}$}

\address{${^1}$KANARI AI , California, USA; 
         ${^2}$Johns Hopkins University, USA; \\
  ${^3}$Qatar Computing Research Institute, Qatar
  }
\copyrightnotice{978-1-6654-7189-3/22/\$31.00~\copyright2023 IEEE}
\begin{document}
\maketitle

\begin{abstract}
The pervasiveness of intra-utterance code-switching (CS) in spoken content requires that speech recognition (ASR) systems handle mixed language. Designing a CS-ASR system has many challenges, mainly due to data scarcity, grammatical structure complexity, and domain mismatch. The most common method for addressing CS is to train an ASR system with the available transcribed CS speech, along with monolingual data. In this work, we propose a zero-shot learning methodology for CS-ASR by augmenting the monolingual data with artificially generating CS text. We based our approach on random lexical replacements and Equivalence Constraint (EC) while exploiting aligned translation pairs to generate random and grammatically valid CS content. Our empirical results show a $65.5$\% relative reduction in language model perplexity, and $7.7$\% in ASR  WER on two ecologically valid CS test sets. The human evaluation of the generated text using EC suggests that more than $80$\% is of adequate quality. 

% Finally, the human evaluation of the generated text suggests that more than 80\% is of adequate quality. 

%%new abstract

% The pervasiveness of intra-utterance code-switching (CS) in spoken content requires that speech recognition (ASR) systems handle mixed language input. Designing a CS-ASR system has many challenges, mainly due to the data scarcity, grammatical structure complexity and mismatch along with unbalanced language use distribution. The most common method for addressing CS phenomena is to train an ASR system with available transcribed CS speech, along with monolingual data from the involved languages.
% In this work, we propose a methodology to augment the monolingual data by artificially  generating Arabic-English CS text to improve different speech modules. We based our approach on Equivalence Constraint theory while exploiting aligned translation pairs, to generate grammatically valid CS content. 
% To 
% In addition, we propose phone similarity edit distance (PSD) for ASR evaluation in the context of CS where the bilingual predictions are mapped into a shared phone space and the substitution cost is scaled by the dissimilarity between the phones.
% % Our empirical results show a 30\% relative reduction in perplexity by adding the synthetic text, and 2\% in WER, on two ecological and noisy CS test sets. 
% Finally, the human evaluation of the generated text suggests that more than 80\% is of adequate quality.
\end{abstract}
\section{Introduction}
\label{sec:intro}

Code-switching (CS) is a prevalent phenomenon in multi-cultural and multi-lingual societies due to the advent of globalization and as a remnant of colonialism. CS, wherein speakers alter between two or more languages during spoken discourse, is now receiving the attention of automatic speech recognition (ASR) researchers, making them address and model mixed-language input to ASR systems. Efforts have been made to design CS-ASR for a variety of language pairs, including Mandarin-English \cite{li2013improved}, Hindi-English \cite{sreeram2020exploration}, French-Arabic \cite{amazouz2017addressing}, Arabic-English \cite{ali2021arabic, hamed2021investigations} and English-French-dialectal Arabic \cite{chowdhury2021onemodel}. Some studies also discuss the complexities of, and the need for, CS ASR between dialects of a language \cite{chowdhury2020effects}. Despite the aforementioned efforts, CS ASR still faces challenges due to the scarcity of transcribed resources, with skewed coverage of languages, dialects and domain mismatch. 

%Therefore,to reduce the effect of the domain mismatch and the data scarcity in speech recognition task, we propose to augment the monolingual text with artificially generated CS data. For this task, we choose Arabic language which, in addition to rich morphological properties, has more than 20 mutually-incomprehensible dialects, with modern standard Arabic (MSA) being the only standardised form \cite{ali2021connecting}. We suggest, however, that the proposed methods should generalized to other languages, e.g. Chinese dialects and Mandarin, subject to availability of the corresponding monolingual resources. 
We propose to augment the monolingual text with artificially generated CS data to reduce the effect of the domain mismatch and the data scarcity in ASR. In our experiments, we choose the Arabic language which is a morphologically complex language with more than 20 mutually-incomprehensible dialects, with modern standard Arabic (MSA) being the only standardized form \cite{ali2021connecting}. The proposed methods should be generalized to other languages, e.g. Chinese dialects and Mandarin, subject to the availability of the corresponding monolingual resources.

% The end-to-end (E2E) systems have gained more popularity recently over conventional hybrid systems. E2E outperformed modular systems in modeling monolingual and multilingual systems \cite{toshniwal2018multilingual, datta2020language,chowdhury2021towards}. This can be owed to the fact that E2E optimizes all parts of the network for the overall word error rate (WER). Researchers in \cite{luo2018towards,shan2019investigating} proposed using additional language identification task on top of connectionist temporal classification (CTC) Attention (CTC-Attention) \cite{kim2017joint} architecture to detect the CS point in English-Mandarin speech. In \cite{sreeram2020exploration}, authors modeled limited Hindi-English CS using E2E attention with context-dependent target to word transduction, factorized language model with part-of-speech (POS) tagging and CS identification, they proposed textual features to enhance the context modeling in CS. In \cite{zhou2020multi}, authors proposed transformer-based architecture with two symmetric language-specific encoders to capture the individual language attributes for Mandarin-English CS, whereas in \cite{chowdhury2021towards}, the authors utilized multilingual strategy to model CS along with dialectal ASR. 

% code switching data generation related work
There have been attempts to explain the grammatical constraints on CS based on Embedded-Matrix theory \cite{myers199511}, Equivalence Constraint \cite{sankoff1998formal} and Functional Head Constraint \cite{belazi1994code}.
In \cite{pratapa2018language}, authors proposed creation of grammatically valid artificial CS data based on the Equivalence Constraint Theory (EC), since it explains a range of interesting CS patterns beyond lexical substitution and is suitable for computational modeling. On the other hand, \cite{winata2019code} proposed to model CS with a sequence-to-sequence recurrent neural network (RNN) with attention, which learns when to switch between and copy words from parallel sentences. 
% However, this approach still requires having considerable amount of real CS data for adaptation.
In \cite{qin2020cosda} researchers proposed generating multi-lingual CS data using mBert alignments with random replacements and then fine-tuning mBert on that data. The proposed approach showed significant improvements in five classification NLP tasks. Although the aforementioned approaches for CS text augmentation provided substantial improvements in downstream NLP tasks, it is not clear if they will help in speech recognition.

In this work, we investigate the effectiveness of different CS text generation approaches to improve ASR performance in a zero-shot learning scenario. 
% A zero-shot learning scenario is a replication of a real-world situation where the data is expected to come from new speakers with different CS styles and different dialect. 
To mitigate the bias to a specific domain during the evaluation, we collect multi-dialectal Arabic-English CS test sets from different domains (Sports, Education, and Interviews) and different dialects (Levantine, Egyptian, Gulf, and Moroccan). In this study, we consider linguistic-based (Equivalence constraint) and lexicon-based (Random) CS generation approaches to answer the following two research questions in the context of ASR performance: $1$) Is the knowledge of the number of switching points important? $2$) Does linguistically motivated CS generation provides improvement over random lexical replacements? 
In addition, we propose a morphologically enhanced pipeline to generate realistic dialectal Arabic-English CS text.
First, we build parallel data by translating original Arabic content into English. Later, we generate the CS content by mixing the pair of parallel sentences guided by the data alignments and different sampling techniques. Our method can be applied without the need for any CS speech data. 

The Arabic language is a morphologically complex language with a high degree of affixation\footnote{A single word could represent multiple tokens. For example, the Arabic segment "\<wsyzrعwnhA fy Hqwlhm>"  ("And they will plant it in their fields") map the first Arabic word to five English tokens, the last word represent the last two. Such $1$-to-n mapping makes it difficult to build natural CS data.}   
and derivation -- it is very challenging to obtain 
accurate alignments with the corresponding English translation. %Moreover ..
% \textcolor{red}{In addition, it is very challenging to obtain the parallel pairs for two languages from sparse dialectal spontaneous speech ==> move elsewhere!} . 
%To address the alignment challenge, 
Thus, we propose to segment the Arabic text into morphemes. The segmentation allows aligning single morphemes with their English translations~\cite{lee-2004-morphological}. 
% We use the Farasa~\cite{abdelali2016farasa} Arabic segmenter for our task. %To address the parallel pairs \textcolor{blue}{@Abdelali.. a little bit about translation we used} \textcolor{red}{Should MT come before? We build parallel text and after we align??}. 
% We show that our novel pipeline is an effective method for realistic CS generation and substantially improves over the baseline ASR and LM models.
We evaluate the well-formed and acceptable generated sentences through subjective evaluation. We compare the mean opinion score (MOS) of the generated text data with ecological transcribed CS speech data, showing majority of the generated utterance are acceptable according to human judgements. The key contributions of this paper are:
\begin{itemize}
\item Create the largest parallel multi-dialectal conversational Arabic-English text corpus.
\item Develop a novel pipeline for generating Arabic-English CS text.
% \item A new milestone of the Arabic-English code-switching speech recognition performance with Hybrid HMM-DNN model.
%\item Comprehensive subjective analysis and human evaluation of the generated Arabic-English code-switching sentences.
\item Analyze human evaluation of the generated Arabic-English CS sentences.
\item Evaluate the efficacy of the generated dialectal Arabic-English CS data in language modeling and ASR.
%\item Comprehensive objective evaluation showing the efficacy of the generated dialectal Arabic-English code-switching data in language modeling and ASR.
% \item Providing new benchmarks with multi-dialectal code switching dataset. 
\end{itemize}

\section{Generating Code-Switched Language}
\label{sec:format}
We begin by describing the proposed approach for generating synthetic Arabic-English CS based on EC theory. 

\subsection{Equivalence Constraint Theory}\label{ec_theory}
In EC theory, both languages $S1$ and $S2$ are defined by context-free grammars $G1$ and $G2$. Every non-terminal category $c1$ in $G1$ has a corresponding non-terminal category $c2$ in $G2$, and every terminal word $w1$ in $G1$ has a corresponding terminal word $w2$ in $G2$. These assumptions imply that intra-sentential code-mixing can only occur at places where the surface structures of two languages map onto each other, hence implicitly following the grammatical rules of both languages. In this work, we build our approach on top of the EC implementation in the GCM toolkit\footnote{https://github.com/microsoft/CodeMixed-Text-Generator}.

\begin{figure*}[hbt!]
\begin{center}
\vspace{-0.3cm}
\includegraphics[width=12cm,height=3.7cm]{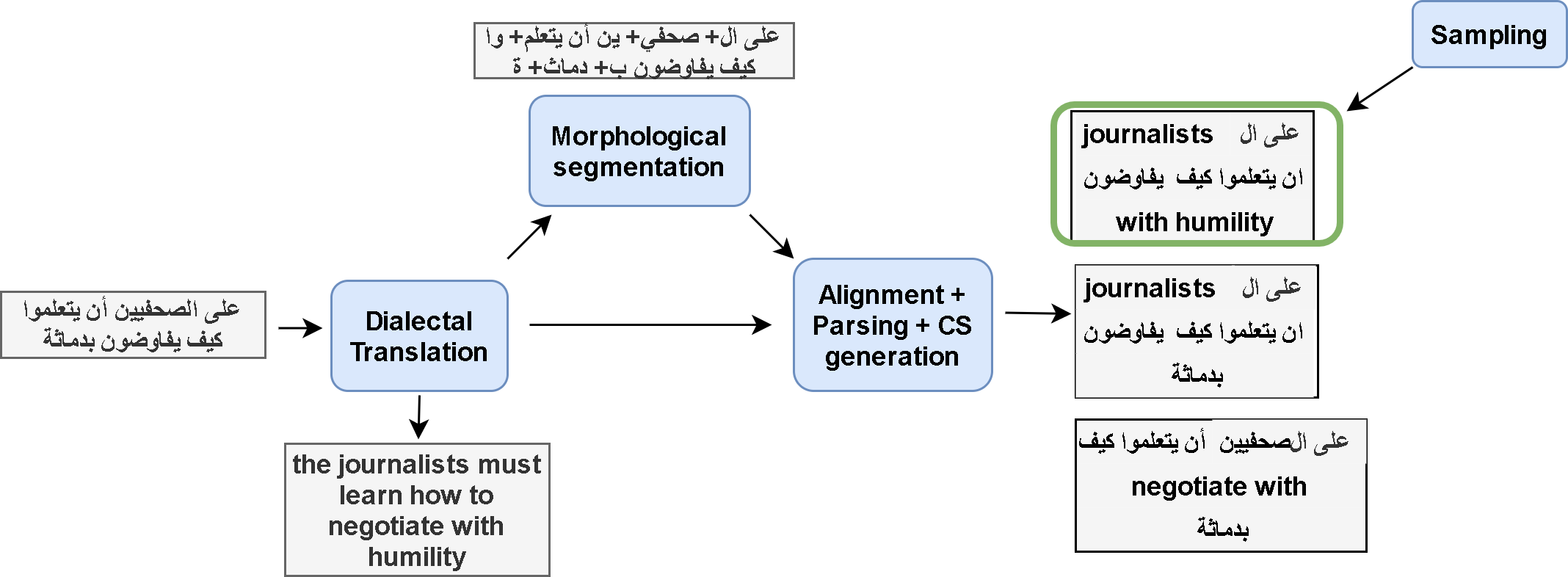}
\vspace{-0.3cm}
\caption{Illustration of the proposed Arabic-English CS generation pipeline.}

\label{fig:cs_pipeline} 
\end{center}
\vspace{-0.5cm}
\end{figure*}

 \subsection{Code-switching Text Generation}
\begin{figure}[hbt!]
 
\begin{center}

\includegraphics[width=8cm,height=4.4cm]{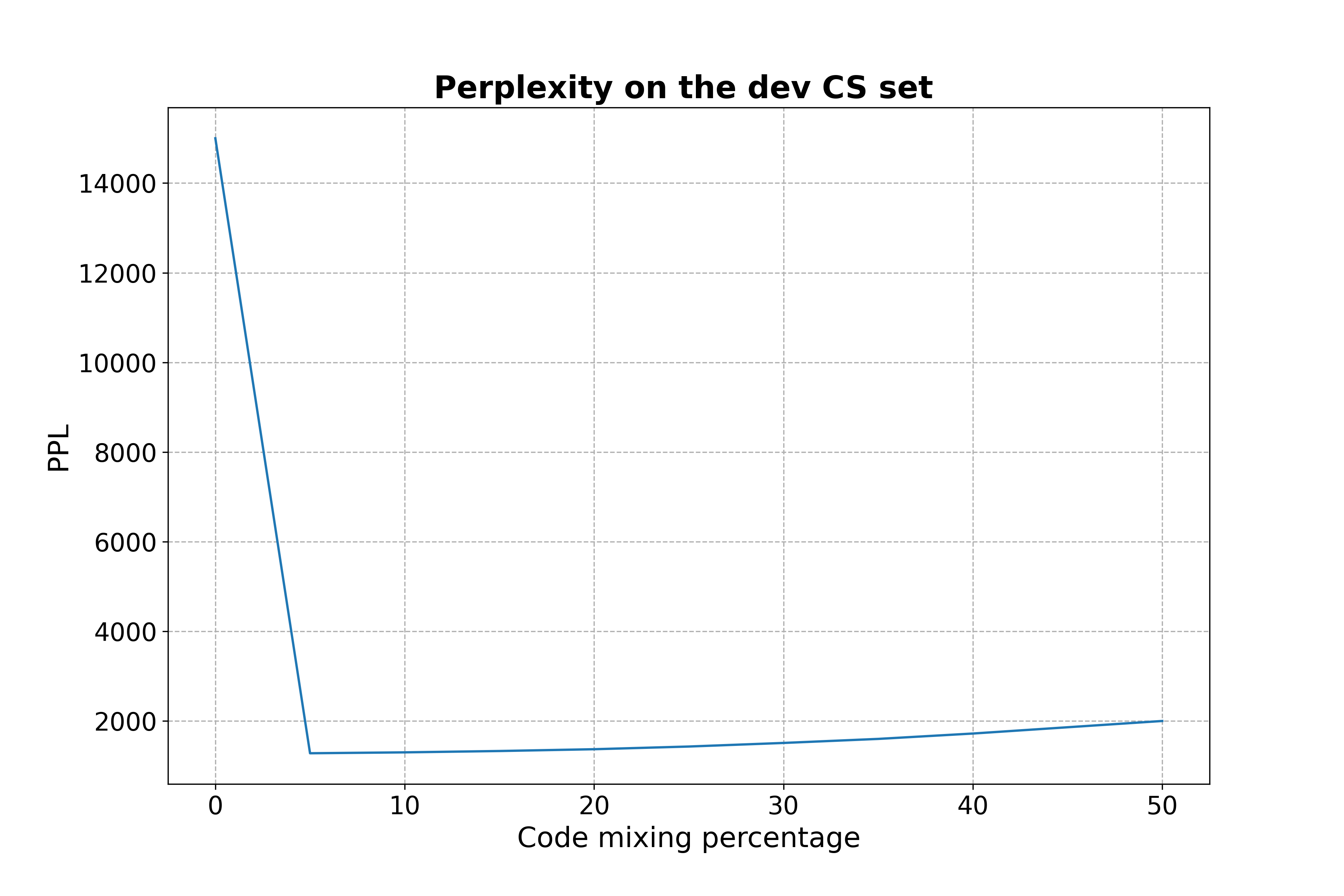}
\vspace{-0.3cm}
\caption{The perplexity of $3$-gram model using generated code-switching text with random lexical replacement  and different code mixing percentages.}

\label{fig:ppl} 
\end{center}
\vspace{-0.2cm}
\end{figure}

\begin{figure*}[h]
\vspace{-0.5cm}
\scalebox{1.0}{
    \subfloat[\label{fig:presegmentedAra}]{
      \includegraphics[width=8cm,height=4.3cm]{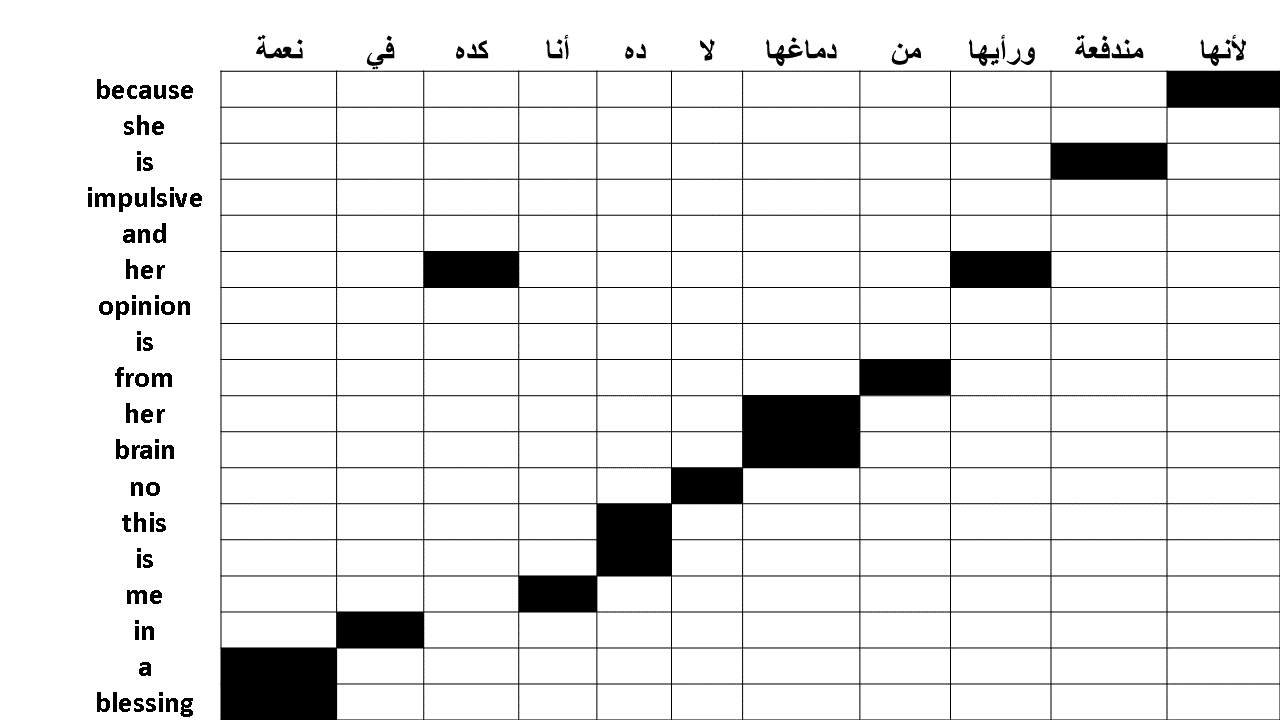}}
\hspace{\fill}
  \subfloat[\label{fig:segmentedAra} ]{
      \includegraphics[ width=8cm,height=4.3cm]{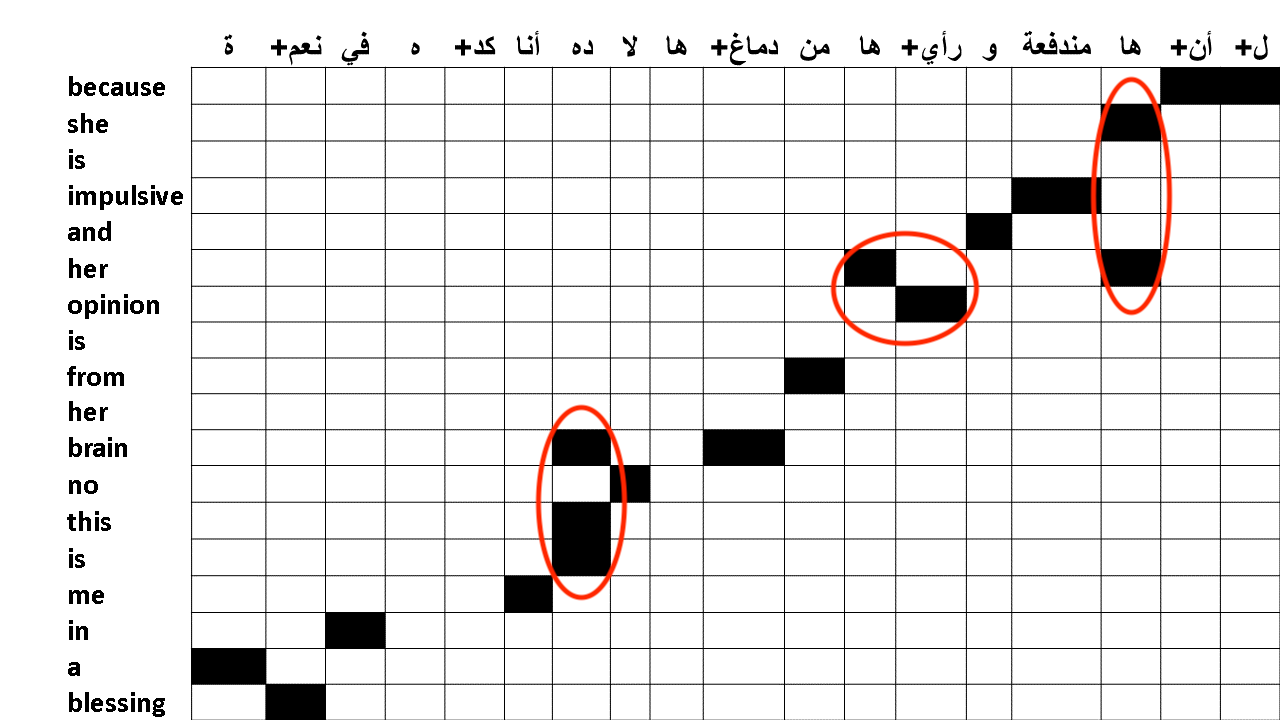}}
\hspace{\fill}
 }
\caption{Illustration of alignments with and without segmentation: (a) Word alignment between Arabic and English without segmentation. (b) Word alignment between Arabic and English with Arabic segmentation. Red circles shows complex construction that were solved after using segmentation.}
\label{fig:segmentation}
\vspace{-0.5cm}
\end{figure*}

 The input to the generation pipeline is a pair of parallel sentences $S1$ and $S2$, along with the word alignments. The $S1$ and $S2$ in our experiments are the English and Arabic languages respectively. The proposed CS generation pipeline includes four major components (also shown in Figure \ref{fig:cs_pipeline}): 
 \begin{enumerate} %[noitemsep,nolistsep]
     \item \textbf{Parallel text translation:} We generate the parallel English text from the Arabic transcription using a public Machine Translation System~\footnote{API access available from \url{https://mt.qcri.org/api}.}, The system is built on transformer-based seq2seq model implemented in OpenNMT~\cite{klein-etal-2017-opennmt}. The Neural translation system is capable of translating Modern Standard Arabic as well as dialectal content~\cite{sajjad-etal-2020-arabench}. It was fine-tuned on a large collection of coarse and fine-grained city-level dialectal data from diverse genres, such as media, chat, religion and travel with varying level of dialectness. Such richness of the system makes it suitable for our task.
     
     \item \textbf{Aligning the two sentences:} To generate word level alignments between the two sentences we use ``fast-align'' \cite{dyer2013simple}, and multilingual Bert (mBert) \cite{dou2021word}. Arabic is agglutinative and morphologically complex language which is difficult to align with Romance and Germanic languages including English. To overcome this challenge, we segmented Arabic words into their stem, prefix(es) and affix(es) using the Farasa \cite{abdelali2016farasa} segmentation. Segmentation has proven to be beneficial in reducing alignment complexity and improving tasks such as Machine Translation~\cite{ataman-federico-2018-evaluation}. Figure~\ref{fig:presegmentedAra} illustrates a complex alignment with several 1-to-many alignments. After using segmentation in Figure~\ref{fig:segmentedAra}, such cases largely disappear. Further,
     the segmentation allows resolving complex constructions that are caused by word re-ordering like ``\<رأي+ ها>'' that is aligned with ``her opinion'' in the reverse order; as well as making it easy to resolve co-references such as  ``\<hA>'' that is mapped to both ``she'' and ``her''.  
     
     \item \textbf{Generating the Parse:} We use Stanford Parser \cite{klein2003accurate} to generate a sentence level constituent parse tree for one of the source languages.  Specifically, we parse the English sentence and use the alignments to generate the equivalent parse tree for the Arabic sentence.
     
     \item \textbf{CS text Generation:} We generate CS text using two approaches: a) random lexical replacements using the alignments from step ($2$) and b) applying EC theory to generate Arabic-English CS text. To examine the effect of changing the percentage of substitutions in the sentence with lexical replacements, we used development set from QASR dataset \cite{mubarak2021qasr} which contains around $16$ hours of CS. Figure \ref{fig:ppl} shows %it can be observed 
     that %the minimum perplexity (PPL) is almost the same along the
     minimum perplexity (PPL) is found to be in the flattened part $5$\%-$30$\%, hence to avoid overfitting, the code switching percentage is selected from around the middle ($20$\%). As for the EC-based CS generation, the high level steps are described as follow:
     \begin{enumerate} 
     
     \item Replace every word in the Arabic parse tree with its English equivalent.
     \item Re-order the child nodes of each internal node in the Arabic tree such that their right-to-left order is similar to the original Arabic language.
     %%% re-wrote by shammur - check
     \item In case of deviation between grammatical structures of the two languages then: 
     \begin{enumerate}%[label=\alph*)]
     \item Replace unaligned English words for any Arabic words with empty strings.
     \item Collapse contiguous word sequences in English, aligned with same Arabic word(s), to a single multi-word node.
     \item Flatten the entire sub-tree, between the above-collapsed nodes and their closest common ancestor, to accommodate the difference.
     \end{enumerate}
    %  \textcolor{red}{Where grammatical structures of two sentences deviates: a) unaligned English words to any Arabic words are replaced with empty strings.
    %  b) Contiguous word sequences in English that are aligned to the same word(s) in Arabic are collapsed into a single multi-word node, and the entire sub-tree between these collapsed nodes and their closest common ancestor is flattened to accommodate this change}.
      \end{enumerate}
      \vspace{-0.5cm}
 \end{enumerate}
% \begin{figure*}[hbt!]
% \begin{center}
% \vspace{-0.3cm}
% \includegraphics[width=0.8\linewidth]{figures/presegmentedArabic.png}
% \vspace{-0.4cm}
% \caption{Word alignment between Arabic and English without segmentation.}
% \label{fig:presegmentedAra} 
% \end{center}
% \end{figure*}

% \begin{figure*}[hbt!]
% \begin{center}
% \vspace{-0.3cm}
% \includegraphics[width=0.8\linewidth]{figures/segmentedArabic.png}
% \vspace{-0.4cm}
% \caption{Word alignment between Arabic and English with Arabic segmentation.}
% \label{fig:segmentedAra} 
% \end{center}
% \end{figure*}

\subsection{Improved Naturalness through Sampling}
In order to generate more natural code-switching (CS) sentences, we experiment with two sampling methods: random sampling and Switch Point Fraction (SPF) sampling. For \textit{random sampling}, we arbitrarily pick a number of CS sentences from the generated data. For \textit{SPF sampling}, we estimate the distribution of the number of switch points in a sentence based on empirical observations mentioned in \cite{chowdhury2021onemodel, mubarak2021qasr} and then we rank the generated CS sentences based on that distribution. We impose the following two constrains to make the generated data more acceptable to a bilingual speaker: $1$) the sentence should start with an Arabic word, and $2$) the number of English words should not exceed $45$\% of the total words in a sentence.
% @amir: how did you got this distribution -- please mention --> fixed by shammur : ask amir to check

 \section{Corpora for Training and Evaluation}
 For empirical analysis of the proposed method, we trained the language models and speech recognition systems. We use several monolingual Arabic\footnote{For code-switching augmentation, we use Arabic monolingual data.} and English datasets. 
 \subsection{Monolingual Data Sets}
 \textbf{MSA and Dialectal Arabic speech data:} We use MSA and multidialectal training data collected from  QASR \cite{mubarak2021qasr}, GALE \cite{olive2011handbook}, MGB3 \cite{ali2017speech}, MGB5 \cite{ali2019mgb}, and an internal 156h Kanari multi-dialectal dataset.
%  Details of train/eval/test split along with genre is presented in Table \ref{tabel1:dataset}.\\
\\
 \textbf{English:} To incorporate a variety of English data, we use the TEDLIUM $3$\footnote{https://openslr.magicdatatech.com/51/} training set and a subset of $300$h of SPGISpeech\footnote{https://datasets.kensho.com/datasets/spgispeech} datasets. 
%  \textcolor{blue}{We note that for acoustic model training we use only the MGB2 \cite{ali2016mgb} which is subset of QASR and TEDLIUM 3.}
 \subsection{Evaluation Code-Switching Data Sets} 
 For comprehensive model evaluation, we use three evaluation sets: monolingual Arabic with multiple dialects (MSA, Egyptian, Gulf, Leventian, North Africa), monolingual English, and two Arabic-English code-switching (CS) sets (ESCWA\footnote{https://arabicspeech.org/escwa} and Kanari multi-dialectal internally collected data).
 \begin{enumerate}
     \item \textbf{Validation set:} for performance evaluation on diverse English and Arabic data, we combine MGB3-test 2h, QASR-test 2h, SPGI-test 1h, and Tedlium3-test 1h.
     \item \textbf{ESCWA-CS} \cite{ali2021arabic}: 2.8h of speech CS data collected over two days of United Nations meetings.
     \item \textbf{Kanari-CS:} 4.8h of  CS data from Levantine, Egyptian, Gulf and Moroccan dialects.
 \end{enumerate}

% \begin{table}[!hbt]
% \centering
% \caption{Data description for train, evaluation and test sets including Arabic (AR) and English (En) languages. The datasets are GALE, MGB3, MGB5, SPGISpeech (SPGI), Tedlium3 (TED), ESCWA-CS and Kanari-CS. 
% }
% %\vspace{-0.3cm}
% \scalebox{1}{
% \begin{tabular}{c|l||c|c|c}
% \hline
% \multicolumn{2}{c||}{Datasets} & Train (hrs) & Eval (hrs) & Test (hrs) \\ 
% \hline
% \multirow{1}{*}{\textbf{En}} &  TED/SPGI  & 452/300 & 1/1 & - \\
% \hline
% \multirow{4}{*}{\textbf{Ar}} & QASR & 1897 & 2 & - \\
%  & MGB3+Dev & \begin{tabular}[c]{@{}c@{}}  20.5\end{tabular} & 2& - \\ %\tablefootnote{4 hours of speech with multiple reference}
%  & MGB5+Dev & \begin{tabular}[c]{@{}c@{}} 48.39\end{tabular} & - & - \\
%  & GALE & 951.4 & - & - \\
%  & Kanari & 156 & - & - \\
%  %\tablefootnote{10.2 hours of speech with multiple reference}
% \hline
% \multirow{3}{*}{\textbf{CS}} & ESCWA-CS & - & - & \begin{tabular}[c]{@{}c@{}}2.8 \end{tabular} \\
%  & Kanari-CS & - & - & \begin{tabular}[c]{@{}c@{}}4.8 \end{tabular} \\
% \hline\hline
% \multicolumn{2}{c|}{TOTAL} & 3825.3 & 6 & 7.6 \\\hline
% \end{tabular}
% }
% \label{tabel1:dataset}
% \end{table}

\begin{table} [!ht]
\centering
\vspace{-0.2cm}
\caption{Mean Opinion Score (MOS) rating scale.}
\vspace{-0.2cm}
\scalebox{0.7}{

\vspace{-0.5cm}
\begin{tabular}{c|c|l}
\textbf{Scores}                     & \multicolumn{1}{c}{\textbf{Labels}} & \multicolumn{1}{c}{\textbf{Definition}}                                                                                                                    \\ 
\hline\hline
\rowcolor[rgb]{1,0.741,0.741} 1     & Very Strange                        & \begin{tabular}[c]{@{}>{\cellcolor[rgb]{1,0.741,0.741}}l@{}}The sentence is never going to be used \\by a human speaker\end{tabular}                       \\
\rowcolor[rgb]{1,0.741,0.741} 2     & Strange                             & \begin{tabular}[c]{@{}>{\cellcolor[rgb]{1,0.741,0.741}}l@{}}The sentence is unlikely to~be used\\by a human speaker\end{tabular}                           \\ 
\hline\hline
\rowcolor[rgb]{0.604,0.741,0.604} 3 & Not very Natural                    & \begin{tabular}[c]{@{}>{\cellcolor[rgb]{0.604,0.741,0.604}}l@{}}The sentence is probably not very natural \\but can be used by human speaker\end{tabular}  \\
\rowcolor[rgb]{0.604,0.741,0.604} 4 & Quite Natural~                      & The sentence can used by human speaker                                                                                                                     \\
\rowcolor[rgb]{0.604,0.741,0.604} 5 & Perfectly Natural~                  & The sentence is definitely used by human speaker                                                                                                          
\end{tabular}
}
\label{tabel2:mos_scale}
\vspace{-0.5cm}
\end{table}

\section{Evaluation of Model Performance}
\label{sec:pagestyle}
We perform objective and subjective evaluations of the quality of the generated CS text. For objective evaluation, we measure the efficacy of the data in language modeling and speech recognition. For subjective evaluation, we asked bilingual annotators to rate the naturalness/acceptability of the utterances.

% . We use LM and ASR for objective evaluation and human score for subjective evaluation. For the LM, we use sri \cite{SRI}, for ASR, we use Kaldi and finally for subjective evaluation, we use figure-8 for crowd sourcing.
\subsection{Language Modeling Evaluation}
We assess the quality of the generated text and its efficacy in handling CS in language modeling (LM) for both $n$-gram and neural LMs.  We build a standard trigram LM using Kneser-Ney smoothing using SRILLM toolkit~\cite{stolcke2002srilm}. Moreover, we train Long short-term memory (LSTM) with $4$ layers, $1,024$ output units and Stochastic gradient descent (SGD) using ESPNet toolkit \cite{watanabe2018espnet} for $20$ epoch.
% We evaluate the efficacy via perplexity (PPL).
%TDNN-LSTM LM using the Kaldi-RNNLM toolkit \cite{xu2018neural} for $5$ epochs. 
 %and out-of-vocabulary (OOV) rates on real CS text.
%traditional $n$-grams (n=3) and recurrent neural LMs. For the  \textcolor{red}{For $n$-grams, we .. @Ahmed} For RNNLMs, we trained neural LMs: TDNN-LSTM \cite{peddinti2017low, povey2018time} using Kaldi-RNNLM toolkit \cite{xu2018neural}. 
% As for evaluating the advantage of generated CS data to imporove PPL in

\subsection{Speech Recognition Evaluation}
Our ASR system uses a hybrid HMM-DNN ASR architecture based on the weighted finite-state transducers (WFSTs) outlined in~\cite{mohri2009weighted}. The training, development, and testing are the same as the Arabic MGB-$2$~\cite{ali2016mgb} and the English TED-LIUM$3$~\cite{hernandez2018ted} tasks. %For more details, refer to Table \ref{tab:asr_data}.
For the hybrid ASR, we trained a Time Delay Neural Network (TDNN) \cite{peddinti2015time} using sequence discriminative training with the LF-MMI objective \cite{povey2016purely} with the alignments from a context-dependent Gaussian mixture model-hidden Markov model (GMM-HMM). The input to the TDNN is composed of $40$-dimensional high-resolution MFCC extracted from $25$ ms frames and $10$ ms shift along with $100$-dimensional i-vectors computed from $1,500$ ms. Five consecutive MFCC vectors and the chunk i-vector are concatenated, forming a $300$-dimensional features vector for each frame. We propose a multilingual architecture that merges all graphemes from multiple languages, keeping the language identity at the grapheme level. A multilingual \textit{n}-gram language model is learned over the transcription for all the languages along with the augmented data.

%%
% Here are some objective evaluations for the data:
% \begin{table}[]
% \centering
% \caption{WER PP} 
% \scalebox{0.9}{
% \begin{tabular}{c|cc|cc|cc}
% \hline
% Data     & \multicolumn{2}{c}{WER in \%} & \multicolumn{2}{c}{Perplexity in 1000} & \multicolumn{2}{c}{number of OOV words} \\
% \hline
% & Kanari&ESCWA & Kanari&ESCWA & Kanari& ESCWA \\
% & & & & & 20,902& 37,416  \\
% \hline
% Baseline & 50.3&47.7 & 4,173&3,216 & 684&814           \\
% Random   & 50.0& 47.5 & 4,071&3,228 & 682&810           \\
% Farasa   & 49.5&47.2 &  3,719&3,117 & 674&782         
% \end{tabular}
% }
% \label{tabel3:wer_pp}
% \end{table}

%%

\subsection{Human Evaluation}
For quality assessment of the generated CS data in terms of naturalness, we designed several crowdsourcing tasks using Amazon Mechanical Turk (MTurk).\footnote{\url{http://mturk.com}}
The tasks aim to rate the utterances' acceptability in terms of the five quality categories ($1$-$5$) (see Table \ref{tabel2:mos_scale}). We consider the green 3+ categories as acceptable sentences by annotators.
The crowdsourcing task is performed using $3$ sets: ($1$) $1$K Generated CS data\footnote{Using equivalence constraints, Farasa segmentation and SPF sampling.}; ($2$) $\approx$$900$ CS data, generated by random lexical replacements; and ($3$) $\approx$$1.9$K utterances from Kanari-CS evaluation dataset containing natural CS. For each task, we collected $3$ judgements per utterances.\footnote{With a cost of $2$ cent per judgment} To ensure the reliability of the annotator, they have to pass an assessment test with a minimum of $80$\% score. The qualifying task includes answering a series of multiple-choice questions - designed by experts to reflect the annotators’ language proficiency and understanding of the questions. A total of $32$ annotators participated in the evaluation. We deliberately put constraints, in the experimental design, such that each evaluator can not annotate more than $15$\% of the utterance from each data set. This ensures that there is no implicit bias encoded in the decision that can influence the reported results.
Using the three judgements, per utterance, we then calculated the mean opinion score (MOS) by averaging the judgment scores. %We in addition, also calculated individual utterance-wise and overall standard deviation between the judgement. 
\section{Empirical Results and Discussion}
\label{sec:typestyle}
\subsection{Objective Evaluation:}
% shammur version
The perplexity (PPL) of the $n$-grams and RNNLM are presented in Table \ref{tabel3:ppl} and WER for the hybrid ASR with $n$-gram LM is reported in Table \ref{tabel3:wer}. 
For the $n$-grams, we observe a significant drop in PPL ($18.6$\% and $22$\% on Kanari and ESCWA data sets respectively) when adding multi-dialectal parallel Arabic-English (Mono) text to the LM training. 
% Adding Amir's
Adding synthetic CS based on word level alignments with equivalence constraint and random sampling improves the PPL further by $3.1$\%-$3.5$\% in relative gain. Using Farasa segmentation with switching point factor (SPF) improves the PPL by $4$\%-$7$\% compared to word level alignment, and by $2.5$\% compared to random sampling. The best PPL is achieved with Farasa segmentation and random lexical replacements with an overall relative gain of $55.5$\% and $65.5$\% on Kanari and ESCWA respectively compared to the baseline.  

For the RNNLMs with fixed BPE tokenizer of size $1k$, we observe a significant reduction in PPL with respect to the monolingual LM (Baselines and Mono) after adding the augmented CS data. This shows that the model is benefiting more from the (synthetic) CS data than increasing the size of monolingual training data. This behavior can be attributed to the fact that subword-based LM can deal with out-of-vocabulary word and segmentation problems more effectively than a word-based n-gram model, thus reducing these factors' influence on the PPL changes. One can notice that the difference is negligible in PPL between  EC-based CS generation techniques (random and SPF). This observation is aligned with the $n$-grams LM results. Finally, random lexical replacements perform significantly better than the EC-based approach.  
% Such pattern is also observed when we used the RNNLM model with an end-to-end ASR system.\footnote{For brevity, we are just presenting results with hybrid ASR results.}

A similar pattern can be observed from the HMM-TDNN performance with LM re-scoring. Table \ref{tabel3:wer} presents the WER on two different test sets in the three experimental settings.  However, WER results indicate that using SPF sampling provides almost no improvements over random sampling. The maximum improvement in WER is obtained with Farasa segmentation and random lexical replacements with a relative gain of $7.7$\% and $4$\% compared to the baseline. We test the significance in the WER improvements using Matched-Pair Sentence Segment Word Error (MAPSSWE) introduced by \cite{gillick1989some}, with a significance level of p=$5$\%. We found that the highest p-value is $0.002$\% as shown in Table \ref{table:st_test}, which is lower than the specified significance level $5$\%. Hence, we reject the null hypothesis and conclude that there is sufficient evidence that the differences in the results are statistically significant.
% \textcolor{red}{For LM and ASR evaluation we report the perplexity and the word error rate (WER) along with the number of insertions deletions and substitutions. For the baseline we use the MGB2+Tedlium data. The perplexity in Table \ref{tabel3:ppl} shows a relative gain of 18.6\% and 22\% on Kanari and ESCWA data-sets respectively when adding multi-dialectal parallel Arabic-English text to the LM training. Adding synthetic CS based on word level alignments with equivalence constraint and random sampling improves the perplexity further by 3.1\%-3.5\% in relative gain. Using Farasa segmentation with switching point factor (SPF) improves the perplexity by 4\%-7\% compared to word level alignment and by 2.5\% compared to random sampling. Finally the best PPL is achieved with Farasa segmentation and random lexical replacements with overall relative gain of 55.5\% and 65.5\% on Kanari and ESCWA respectively compared to the baseline. A similar pattern can be observed from ASR performance with LM re-scoring. However WER results indicates that using SPF sampling provides almost no improvements over random sampling. The maximum improvement in WER is obtained with Farasa segmentation and lexical replacements with relative gain of 7.7\% and 4\% compared to the baseline. 
% }% using Farasa alignments with both sampling methods. Our results indicates the importance of considering the code-alteration distribution (i.e., SPF) with a relative gain of $\approx9\%$ and $\approx3.5\%$ in PPL wrt random sampling in $n$-grams for both the test sets. A similar pattern is seen in RNNLMs. \\

\begin{table}[!htb]
\centering
\caption{Perplexity for $n$-gram and RNNLM on Kanari and ESCWA test sets. MGB$2$+Tedlium$3$ transcription (Baseline), collected Arabic English parallel text (Mono); Fast Align (FA); Equivalence Constraint (EC); Lexical Replacements (LR); Switch Point Fraction (SPF); random sampling with word alignment (Random); random sampling with Farasa segmentation alignment (Farasa+Random).}
\vspace{-0.2cm}
\scalebox{0.67}{
\begin{tabular}{l|rr||rr||r}
 \textit{Perplexity}    & \multicolumn{2}{c||}{Kanari}  & \multicolumn{2}{c||}{ESCWA} & \\ 
\hline\hline
\#Tokens & \multicolumn{2}{c||}{20,902} & \multicolumn{2}{c||}{37,416}& \\ 
\hline\hline
                      & \multicolumn{1}{l}{$n$-gram} & \multicolumn{1}{l||}{LSTM} & \multicolumn{1}{l}{$n$-gram} & LSTM & \multicolumn{1}{|l}{\#sent}  \\ 
\hline\hline
Baseline             &  $5,284$    &    $127$  & $5,565$    &   $87$  &      $784$K  \\  \hline 
\MyIndent+Mono (Ar-En)  & $4,456$     &    $179$   &         $4,565$           &  $118$   &   $3.108$M                \\  \hline
% &&&&&&&&\\  \hline
\MyIndent \MyIndent +EC+Random (FA)   &   $4,318$     &    $50$    &        $4,412$     & $46$    &   $3.677$M          \\ \hline
\MyIndent \MyIndent +EC+Farasa+Random (FA)    &   $4,206$ &  $47$ &      $4,379$       & $46$    &   $3.677$M   \\ \hline
\MyIndent \MyIndent +EC+Farasa+SPF (FA)  &     $4,102$    &     $51$    &         $4,272$       &  $50$  &  $3.677$M   \\ \hline
\MyIndent \MyIndent +EC+Farasa+SPF (mBert)  &     $3,995$    &     $53.8$    &   $4,202$ &  $51$  &  $3.677$M
 \\ \hline
\MyIndent \MyIndent +EC+ Farasa +SPF   & $4,038$ & $59$ &   $4,236$ & $58$ & $4.246$M \\
\MyIndent \MyIndent   \MyIndent \MyIndent    + Random (FA)    &  & & & & \\ \hline

\MyIndent \MyIndent +Farasa+LR (mBert)    & \textbf{$3,398$}& \textbf{$43$}&    \textbf{$3,362$}& \textbf{$45$} & 3.603M\\ \hline
% \MyIndent \MyIndent +Farasa+lexical\_all    & \textbf{3195}  &  & \textbf{3130} & & 4.585M  \\ \hline
\end{tabular}}
\label{tabel3:ppl}
\vspace{-0.2cm}
\end{table}

\begin{table}[!htb]
\centering
\caption{WER with insertions: ins, deletions: del and substitutions: sub on Kanari and Escwa test sets. MGB$2$+Tedlium$3$ transcription (Baseline), (Mono): collected Arabic English parallel text; Fast Align (FA); Lexical Replacements (LR); Equivalence Constraint (EC); Switch Point Fraction (SPF); random sampling with word alignment (Random); random sampling with Farasa segmentation alignment (Farasa+Random).}
\label{tabel3:wer}

\scalebox{0.65}{
\begin{tabular}{l|c|c}
\textit{Hybrid ASR}    & \multicolumn{2}{c}{WER in \% \& [ins, del ,sub] }  \\ 
\hline
\textit{LM Data} & Kanari  & ESCWA     \\ 
\hline
Baseline      & $59.30$ [$532$, $9,094$ ,$15,282$]   & $49.25$ [$419$, $3,668$, $6,086$]           \\ \hline
\MyIndent + Mono (Ar-En) & $56.27$ [$587$, $8,253$, $14,796$] & $47.80$ [$365$, $3,961$, $5,549$] \\ \hline
\MyIndent \MyIndent +EC + Random (FA )& $56.18$ [$596$, $8,219$, $14,782$]  & $47.79$ [$450$, $3,418$, $6,005$]    \\ \hline
\MyIndent \MyIndent +EC + Farasa + Random (FA)  & $55.89$ [$603$, $8,093$, $14,778$]   & $47.62$ [$464$, $3,372$, $6,000$] \\ \hline
\MyIndent \MyIndent +EC + Farasa +SPF (FA)  &  $55.87$ [$590$, $8,107$, $14,769$]   & $47.64$ [$370$, $3,927$, $5,545$]   \\  \hline
\MyIndent \MyIndent +EC+Farasa+SPF (mBert)  &  $55.85$ [$608$, $8,119$, $14,733$]   & $47.54$ [$451$, $3,373$, $5,996$] \\\hline
\MyIndent \MyIndent +EC+Farasa+SPF    & $55.81$ [$603$, $8,075$, $14,763$] & $47.67$ [$373$, $3,940$, $5,535$]                 \\ 
\MyIndent \MyIndent \MyIndent\MyIndent+Random (FA) & & \\ \hline
\MyIndent \MyIndent +Farasa+LR (mBert) & \textbf{$55.04$} [$618$, $7,803$, $14,698$] & \textbf{$47.28$} [$491$, $3,302$, $5,973$]     \\ \hline
%\MyIndent \MyIndent +Farasa+lexical\_all& &  \\\hline
\end{tabular}
}
\vspace{-0.5cm}
\end{table}

\begin{table}[!htb]
\centering
\caption{MAPSSWE p-values between the baseline (B), baseline+mono (B+M), and baseline+Mono+CS with lexical replacements (B+M+LR) on Kanari and Escwa sets } % u: represents total utterances used.}
% \scalebox{0.8}{
\vspace{-0.3cm}
\begin{tabular}{c|c c|c c}
& \multicolumn{2}{c|}{ESCWA} & \multicolumn{2}{c}{Kanari} \\ 
\hline
\hline
     & B & B+M & B & B+M \\
 B+M & $< 0.001$ & - &$< 0.001$ & - \\
 B+M+LR &$< 0.001$ & $0.002$ & $< 0.001$ & $< 0.001$
 \\
 \hline
\end{tabular}
% }
\label{table:st_test}
\vspace{-0.3cm}
\end{table}
\begin{table*}[!htb]
\centering

\caption{Reported percentage of human judgement scores data falls under the MOS value range for the human-transcribed CS data (Kanari-CS), generated CS data using Equivalence Constrain (EC:SPF) and random lexical replacement (RLR). The $x<=*<y$ represent the MOS value range. } % u: represents total

% \caption{Reported percentage of average MOS score (collected from 3 judgments)
% data falls under the MOS value range for the human-transcribed CS data (Kanari-CS), generated CS data using Equivalence Constrain (EC:SPF) and random lexical replacement (RLR). The $x<=*<y$ represent the MOS value range. The MOS value  } % u: represents total utterances used.}
% \scalebox{0.8}{
\vspace{-0.25cm}
\begin{tabular}{c|cc|cc|cc}
\textbf{MOS} & \multicolumn{2}{c|}{EC:SPF (\# $1,170$)} & \multicolumn{2}{c|}{RLR (\# $900$)} & \multicolumn{2}{c}{Kanari-CS (\# $1,921$)} \\ \cline{1-5} \cline{6-7} 
$1<= *<2$                      & $1.20$\%     & \multirow{2}{*}{$16.15$\%}    & $9.78$\%     & \multirow{2}{*}{$74.67$\%}   & $4.22$\%     & \multirow{2}{*}{$41.96$\%}   \\ 
$2<= *<3$                      & $14.96$\%    &                             & $64.89$\%   &                          & $37.74$\%    &                            \\ \cline{1-7} %\cline{6-7} 
$3<= *<4$                      & $54.36$\%    & \multirow{2}{*}{$83.85$\%}    & $22.11$\%    & \multirow{2}{*}{25.33\%}   & $43.68$\%    & \multirow{2}{*}{$58.04$\%}   \\
$4<=* <=5$                     & $29.49$\%    &                             & $3.22$\%    &                          & $14.37$\%    &                           
\end{tabular}
% }
\label{tabel3:mos_rslt}
\vspace{-0.3cm}
\end{table*}

\subsection{Subjective Evaluation:}
The percentage of average judgment scores for each quality category are presented in Table \ref{tabel3:mos_rslt}.
From MOS, we observe that $\approx84\%$ of the generated data with equivalence constraint and switch point factor is acceptable, while a random replacement is only $25$\% acceptable to human judges. Hence, reflecting the importance of the proposed pipeline for enriching CS data. Our result also suggests that generated CS data is cleaner than natural CS transcription, which contains overlapping speech, disfluency, and repetition among others. On the other hand, even though random replacement is less acceptable by humans, it helps the LM/ASR to see and learn more about CS as an effective augmentation technique. We further discuss this in the following Section.  

\subsection{Key Observations and Discussion}
Both LM, and ASR with LM re-scoring results suggest that random lexical replacements provide the best performance on code-switching (CS) when testing on a new domain. We think that this is because CS, in general, is affected by different factors including the dialect, the topic, the social and educational status, the emotional state, the speaker's speech styles, and the proficiency in the two languages. Hence, CS in a new domain is an unpredictable phenomenon and in practice can be modeled as a random process. Furthermore, the ASR results show that the CS generation with the expected number of switching points (SPF) provides marginal improvements compared to random replacements. We think this is mainly because in a zero-shot learning scenario, in addition to SPF, the model needs information about the position where the switching is expected to happen.

%Human evaluation of the synthetic text indicates the potential of the proposed method to generate human-approved CS data. The addition of CS-augmented data to LM shows a significant improvement in perplexity when combining Farasa with any text generation settings. 

Human evaluation of the synthetic text shows that our method can generate natural CS augmented text. Furthermore, adding the CS data show significant improvement in PP and WER when combining Farasa with any text generation settings.

%when combining Farasa and SPF settings. 
% It may be noted that such gain is not due to changes in OOV rate. As seen from the Table \ref{tabel3:ppl}, CS-augmented data does not help much with unknown words above the monolingual sets.
Finally, a challenge in evaluating CS ASR is the metric itself. As shown in recent CS studies \cite{chowdhury2021onemodel,injySLT2022}, WER is not robust against partial/full transliteration of a correctly recognized word, hence doesn't fully reflect improvement in CS-ASR, as seen in the example below, where the words ``International'' and ``Pharmatech'' have a script mismatch between the reference and ASR output: \\
\begin{center}
\vspace{-0.6cm}
    \img{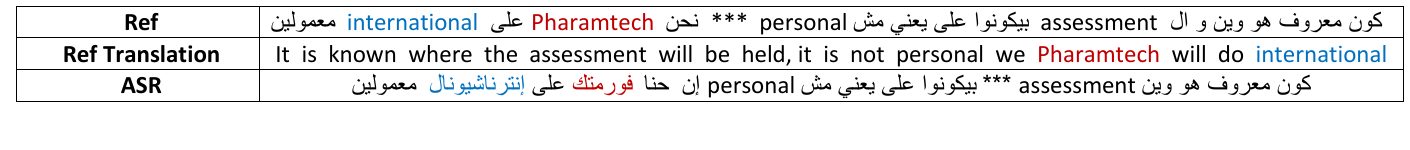}
    \vspace{-0.6cm}
\end{center}
While the above combination of scores confirms the validity of the approach, inspecting sentences with low MOS and/or high perplexity reveals some shortcomings of the proposed approach. Several sentences that were scored lower than $3$ had one common issue: the original sentences were not complete as shown below:
% \begin{figure}[!ht]
% \centering
%  \includegraphics[width=\linewidth]{figures/cs_example.pdf}
%  \vspace{-0.9cm}
%   \caption{ }\label{fig:cs_example}
%   \vspace{-0.3cm}
% \end{figure}
\begin{center}
\vspace{-0.6cm}
    \img{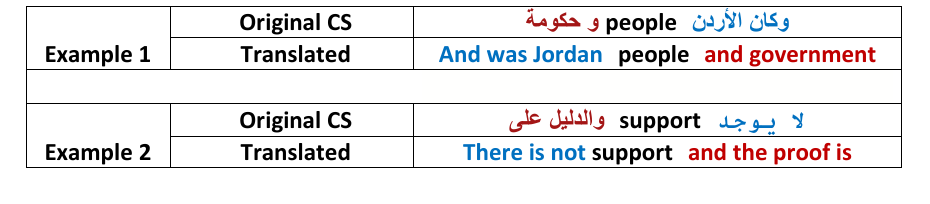}
    \vspace{-0.6cm}
\end{center}
% In speech fragments such as ``\arastr{و حكومة} people \arastr{وكان الأردن}''
% (``{\em And was Jordan, people and government}'')
% and 
% ``is \arastr{والدليل على} support \arastr{لا يوجد}'''
% (``{\em There is not support and the proof is}''),
Though the CS algorithm chose the correct English replacement, the context seems missing or incomplete/broken, leading to a low score by the human judges. 
An additional issue was errors inherent in the MT system.  Incorrect lexical choices create further ambiguity in the resulting CS sentences and result in low MOS. Some instances were a result of the neural MT generating more fluent output than its (disfluent) input, causing misalignment between the original and translated sentences, while impacting the quality of the generated CS. Restricting the generation in the step ($3$) to complete sentences while parsing the sentences would avoid some of the aforementioned problems. Additionally, using other MT systems can help to avoid the low-quality translations that degrade the CS generation.

\section{Conclusion}
%%re-write
In this work, we have proposed a novel pipeline for Arabic-English code-switching text generation to improve speech modules. We show that morphological segmentation is crucial to obtain accurate alignments to synthesize more realistic CS. Through objective evaluation, we show that the best approach in the zero-shot learning scenario is achieved with morphological segmentation and random lexical replacements. The proposed approach substantially improves both LM and ASR by $55.5$\%-$65.5$\% and $4$\%-$7.7$\% relative gain in perplexity and WER respectively on code-switching data from a new domain. Subjective evaluation for the naturalness of the generated CS text indicates that $84$\% of the generated sentences are acceptable to human judges. The proposed method can be generalized for other language pairs, depending on the available monolingual resources.

\section{ACKNOWLEDGMENTS}
The work presented here was carried out during the $2022$ Jelinek Memorial Summer Workshop on Speech and Language Technologies at Johns Hopkins University, which was supported with funding from Amazon, Microsoft and Google.
\\
\bibliographystyle{IEEEbib}
\bibliography{strings,refs}

\begin{thebibliography}{10}

\bibitem{li2013improved}
Y.~Li and P.~Fung,
\newblock ``Improved mixed language speech recognition using asymmetric
  acoustic model and language model with code-switch inversion constraints,''
\newblock in {\em ICASSP}, 2013.

\bibitem{sreeram2020exploration}
G.~Sreeram and R.~Sinha,
\newblock ``Exploration of end-to-end framework for code-switching speech
  recognition task: Challenges and enhancements,''
\newblock {\em IEEE Access}, 2020.

\bibitem{amazouz2017addressing}
D.~Amazouz, M.~Adda-Decker, and L.~Lamel,
\newblock ``Addressing code-switching in {F}rench/{A}lgerian {A}rabic speech,''
\newblock in {\em Interspeech}, 2017.

\bibitem{ali2021arabic}
A.~Ali, S.~Chowdhury, A.~Hussein, and Y.~Hifny,
\newblock ``Arabic code-switching speech recognition using monolingual data,''
\newblock {\em Interspeech 2021}, 2021.

\bibitem{hamed2021investigations}
I.~Hamed, P.~Denisov, C.~Li, M.~Elmahdy, S.~Abdennadher, and N.g Vu,
\newblock ``Investigations on speech recognition systems for low-resource
  dialectal {A}rabic-{E}nglish code-switching speech,''
\newblock {\em Computer Speech \& Language}, p. 101278, 2021.

\bibitem{chowdhury2021onemodel}
S.~Chowdhury, A.~Hussein, A.~Abdelali, and A.~Ali,
\newblock ``Towards one model to rule all: Multilingual strategy for dialectal
  code-switching {A}rabic {A}sr,''
\newblock {\em Interspeech 2021}, 2021.

\bibitem{chowdhury2020effects}
S.~A Chowdhury, Y.~Samih, M.~Eldesouki, and A.~Ali,
\newblock ``Effects of dialectal code-switching on speech modules: A study
  using egyptian {A}rabic broadcast speech,''
\newblock {\em Interspeech}, 2020.

\bibitem{ali2021connecting}
A.~Ali, S.~Chowdhury, M.~Afify, W.~El-Hajj, H.~Hajj, M.~Abbas, A.~Hussein,
  N.~Ghneim, M.~Abushariah, and A.~Alqudah,
\newblock ``Connecting {A}rabs: bridging the gap in dialectal speech
  recognition,''
\newblock {\em Communications of the ACM}, 2021.

\bibitem{myers199511}
C.~Myers-Scotton,
\newblock ``11 a lexically based model of code-switching,''
\newblock {\em One speaker, two languages: Cross-disciplinary perspectives on
  code-switching}, 1995.

\bibitem{sankoff1998formal}
D.~Sankoff,
\newblock ``A formal production-based explanation of the facts of
  code-switching,''
\newblock {\em Bilingualism: language and cognition}, 1998.

\bibitem{belazi1994code}
H.~Belazi, E.~Rubin, and A.~Toribio,
\newblock ``Code switching and x-bar theory: The functional head constraint,''
\newblock {\em Linguistic inquiry}, 1994.

\bibitem{pratapa2018language}
A.~Pratapa, G.~Bhat, M.~Choudhury, S.~Sitaram, S.~Dandapat, and K.~Bali,
\newblock ``Language modeling for code-mixing: The role of linguistic theory
  based synthetic data,''
\newblock in {\em ACL}, 2018.

\bibitem{winata2019code}
G.~Winata, A.~Madotto, C.~Wu, and P.~Fung,
\newblock ``Code-switched language models using neural based synthetic data
  from parallel sentences,''
\newblock {\em arXiv preprint arXiv:1909.08582}, 2019.

\bibitem{qin2020cosda}
Libo Qin, Minheng Ni, Yue Zhang, and Wanxiang Che,
\newblock ``Cosda-ml: Multi-lingual code-switching data augmentation for
  zero-shot cross-lingual nlp,''
\newblock {\em arXiv preprint arXiv:2006.06402}, 2020.

\bibitem{lee-2004-morphological}
Y.~Lee,
\newblock ``Morphological analysis for statistical machine translation,''
\newblock 2004.

\bibitem{klein-etal-2017-opennmt}
G.~Klein, Y.~Kim, Y.~Deng, J.~Senellart, and A.~Rush,
\newblock ``{O}pen{NMT}: Open-source toolkit for neural machine translation,''
\newblock in {\em ACL}, 2017.

\bibitem{sajjad-etal-2020-arabench}
H.~Sajjad, A.~Abdelali, N.~Durrani, and F.~Dalvi,
\newblock ``{A}ra{B}ench: Benchmarking dialectal {A}rabic-{E}nglish machine
  translation,''
\newblock in {\em Proceedings of the 28th International Conference on
  Computational Linguistics}, 2020.

\bibitem{dyer2013simple}
C.~Dyer, V.r Chahuneau, and N.~Smith,
\newblock ``A simple, fast, and effective reparameterization of {IBM} model
  2,''
\newblock in {\em NACL: Human Language Technologies}, 2013.

\bibitem{dou2021word}
Zi-Yi Dou and Graham Neubig,
\newblock ``Word alignment by fine-tuning embeddings on parallel corpora,''
\newblock in {\em Conference of the European Chapter of the Association for
  Computational Linguistics (EACL)}, 2021.

\bibitem{abdelali2016farasa}
A.~Abdelali, K.~Darwish, N.~Durrani, and H.~Mubarak,
\newblock ``Farasa: {A} fast and furious segmenter for {A}rabic,''
\newblock in {\em NACL: Demonstrations}, 2016.

\bibitem{ataman-federico-2018-evaluation}
D.~Ataman and M.~Federico,
\newblock ``An evaluation of two vocabulary reduction methods for neural
  machine translation,''
\newblock in {\em Association for Machine Translation in the {A}mericas}, 2018.

\bibitem{klein2003accurate}
D.~Klein and C.~Manning,
\newblock ``Accurate unlexicalized parsing, in proceedings of the 41st meeting
  of the association for computational linguistics,''
\newblock 2003.

\bibitem{mubarak2021qasr}
H.~Mubarak, A.~Hussein, S.~Chowdhury, and A.~Ali,
\newblock ``{QASR: QCRI A}ljazeera speech resource {A} large scale annotated
  arabic speech corpus,''
\newblock in {\em ACL}, 2021.

\bibitem{olive2011handbook}
J.~Olive, C.~Christianson, and J.~McCary,
\newblock {\em Handbook of natural language processing and machine translation:
  DARPA global autonomous language exploitation},
\newblock Springer Science \& Business Media, 2011.

\bibitem{ali2017speech}
A.~Ali, S.~Vogel, and S.~Renals,
\newblock ``Speech recognition challenge in the wild: {A}rabic {MGB}-3,''
\newblock in {\em ASRU}, 2017.

\bibitem{ali2019mgb}
A.~Ali, S.~Shon, Y.~Samih, H.~Mubarak, A.~Abdelali, J.~Glass, S.~Renals, and
  K.~Choukri,
\newblock ``The {MGB}-5 challenge: Recognition and dialect identification of
  dialectal {A}rabic speech,''
\newblock in {\em ASRU}, 2019.

\bibitem{stolcke2002srilm}
A.~Stolcke,
\newblock ``{SRILM}-an extensible language modeling toolkit,''
\newblock in {\em Seventh international conference on spoken language
  processing}, 2002.

\bibitem{watanabe2018espnet}
Shinji Watanabe, Takaaki Hori, Shigeki Karita, Tomoki Hayashi, Jiro Nishitoba,
  Yuya Unno, Nelson Enrique~Yalta Soplin, Jahn Heymann, Matthew Wiesner, Nanxin
  Chen, et~al.,
\newblock ``Espnet: End-to-end speech processing toolkit,''
\newblock {\em arXiv preprint arXiv:1804.00015}, 2018.

\bibitem{mohri2009weighted}
M.~Mohri,
\newblock ``Weighted automata algorithms,''
\newblock in {\em Handbook of weighted automata}. 2009.

\bibitem{ali2016mgb}
A.~Ali, P.~Bell, J.~Glass, Y.~Messaoui, H.~Mubarak, S.~Renals, and Y.~Zhang,
\newblock ``The {MGB}-2 challenge: {Arabic} multi-dialect broadcast media
  recognition,''
\newblock in {\em SLT}, 2016.

\bibitem{hernandez2018ted}
F.~Hernandez, V.~Nguyen, S.~Ghannay, N.~Tomashenko, and Y.~Est{\`e}ve,
\newblock ``{TED-LIUM} 3: twice as much data and corpus repartition for
  experiments on speaker adaptation,''
\newblock in {\em International Conference on Speech and Computer}, 2018.

\bibitem{peddinti2015time}
V.~Peddinti, D.~Povey, and S.~Khudanpur,
\newblock ``A time delay neural network architecture for efficient modeling of
  long temporal contexts,''
\newblock in {\em Interspeech}, 2015.

\bibitem{povey2016purely}
D.~Povey, V.~Peddinti, D.~Galvez, P.~Ghahremani, V.~Manohar, X.~Na, Y.~Wang,
  and S.~Khudanpur,
\newblock ``Purely sequence-trained neural networks for {ASR} based on
  lattice-free {MMI},''
\newblock in {\em Interspeech}, 2016.

\bibitem{gillick1989some}
Laurence Gillick and Stephen~J Cox,
\newblock ``Some statistical issues in the comparison of speech recognition
  algorithms,''
\newblock in {\em International Conference on Acoustics, Speech, and Signal
  Processing,}. IEEE, 1989, pp. 532--535.

\bibitem{injySLT2022}
I.~Hamed, A.~Hussein, O.~Chellah, S.~Chowdhury, H.~Mubarak, S.~Sitaram,
  N.~Habash, and A.~Ali,
\newblock ``Benchmarking evaluation metrics for code-switching automatic speech
  recognition,''
\newblock in {\em SLT}, 2023.

\end{thebibliography}

\end{document}